\documentclass[11pt]{article}

\usepackage{hyperref}

\usepackage{fullpage,url,hyperref,amssymb,graphicx,listings,xcolor}
\usepackage{rotating}
\usepackage{pifont}
\usepackage{xspace}
\usepackage[T1]{fontenc}
\usepackage[utf8]{inputenc}

\graphicspath{{./figures/}{./}}

\makeatletter
\let\@xobeysp\space
\makeatother

\def\citep#1{\cite{#1}}

\newcommand{\pyBregMan}{{\tt pyBregMan}\xspace}

\def\bbR{\mathbb{R}}
\def\inner#1#2{{\left\langle #1,#2\right\rangle}}

\def\st{{\ :\ }}

\def\GL{\mathrm{GL}}

\def\barF{{\bar{F}}}

\def\bbR{\mathbb{R}}

\makeatletter\newcommand*\bigcdot{\mathpalette\bigcdot@{.5}}
\newcommand*\bigcdot@[2]{\mathbin{\vcenter{\hbox{\scalebox{#2}{$\m@th#1\bullet$}}}}}

\DeclareRobustCommand\onedot{\futurelet\@let@token\bmv@onedotaux}
\def\bmv@onedotaux{\ifx\@let@token.\else.\null\fi\xspace}
\def\eg{\emph{e.g}\onedot} 
\def\ie{\emph{i.e}\onedot} 
 
\def\etc{\emph{etc}\onedot} 
\def\wrt{w.r.t\onedot}

\makeatother

\usepackage{fancyvrb}

\makeatletter
\def\VerbLB{\FV@Command{}{VerbLB}}
\begingroup
\catcode`\^^M=\active%
\gdef\FVC@VerbLB#1{%
  \begingroup%
    \FV@UseKeyValues%
    \FV@FormattingPrep%
    \FV@CatCodes%
    \def^^M{ }%
    \catcode`#1=12%
    \def\@tempa{\def\FancyVerbGetVerb####1####2}%
    \expandafter\@tempa\string#1{\mbox{##2}\endgroup}%
    \FancyVerbGetVerb\FV@EOL}%
\endgroup
\makeatother

\title{\pyBregMan: A Python library for Bregman Manifolds\thanks{\url{https://franknielsen.github.io/pyBregMan/}}}

\date{}

\author{Frank Nielsen\thanks{Sony Computer Science Laboratories Inc., Tokyo,
Japan.} \and Alexander Soen\thanks{Australian National University, Canberra, Australia; RIKEN AIP, Tokyo, Japan.}}

\begin{document}

    \maketitle

\begin{abstract}
A {\em Bregman manifold} is a synonym for a dually flat space in information geometry which admits as a canonical divergence a Bregman divergence. Bregman manifolds are induced by smooth strictly convex functions like the cumulant or partition functions of regular exponential families, the negative entropy of mixture families, or the characteristic functions of regular cones just to list a few such convex Bregman generators. 
We describe the design of \pyBregMan, a library which implements generic operations on Bregman manifolds and instantiate several common Bregman manifolds used in information sciences.
At the core of the library is the notion of Legendre-Fenchel duality inducing a canonical pair of dual potential functions and dual Bregman divergences.
The library also implements the Fisher-Rao manifolds of categorical/multinomial distributions and multivariate normal distributions.
To demonstrate the use of the  \pyBregMan{} kernel manipulating those Bregman and Fisher-Rao manifolds, the library also provides several core algorithms for various applications in statistics, machine learning, information fusion, and so on.
\end{abstract}


    \tableofcontents

\section{Getting started with \pyBregMan}
\label{sec:getting-started}
 
Let us build, step-by-step, a simple example to demonstrate usage of the \pyBregMan library 
(\underline{Py}thon library for \underline{Breg}man \underline{Man}ifolds):\\

\begin{figure}
    \centering
    \includegraphics[width=1.0\textwidth]{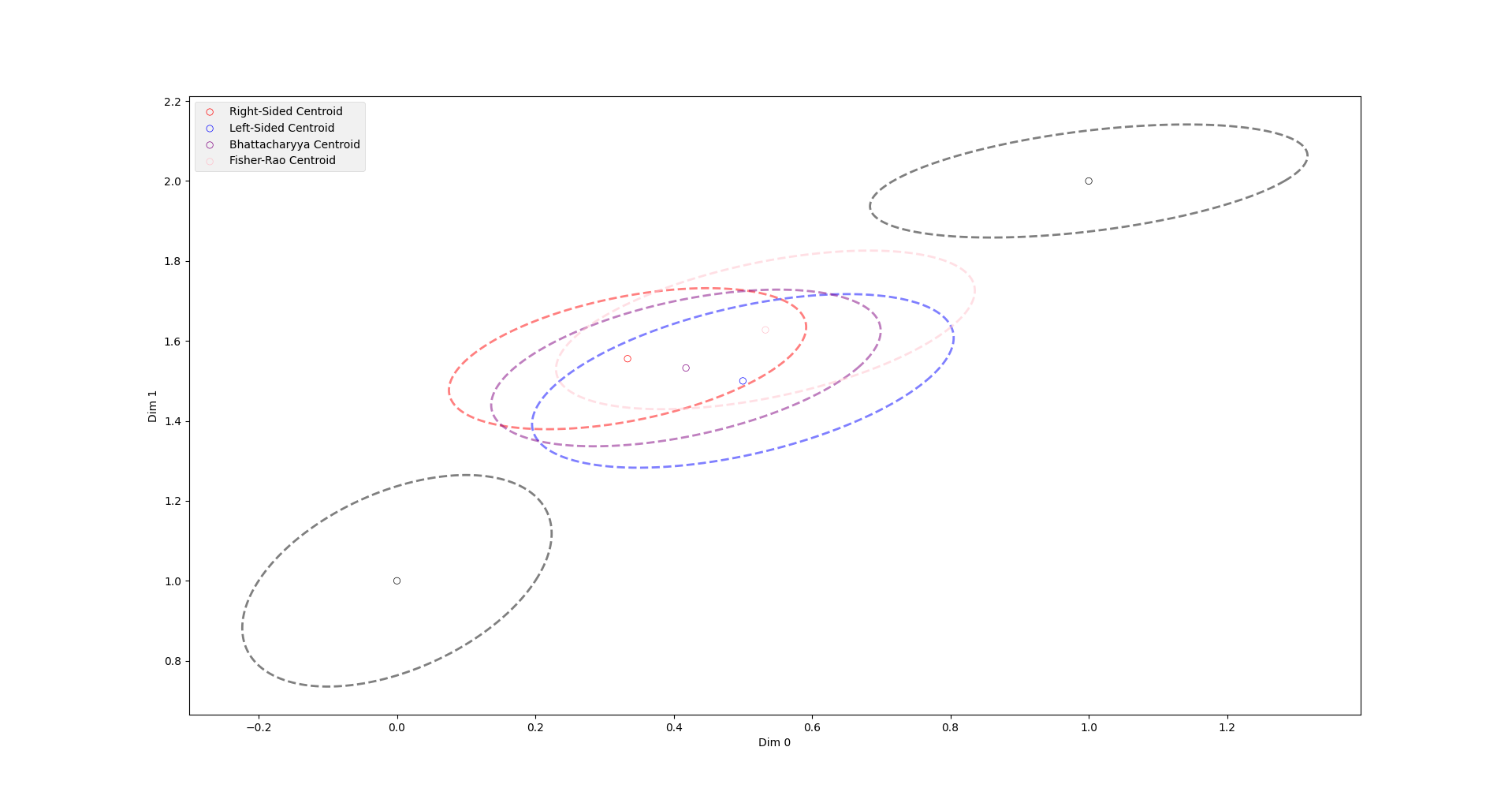}
    \caption{%
        Left-sided Kullback-Leibler, right-sided Kullback-Leibler, Bhattacharyya
        and Fisher-Rao centroids on the bivariate normal manifold.
    }\label{fig:mvncentroids}
\end{figure}

\noindent \ding{172} we create the Bregman and Fisher-Rao manifold  structures for the
family of bivariate normal distributions (namely, an exponential family
manifold and a Fisher-Rao manifold);\\
\noindent \ding{173} we calculate the Kullback-Leibler
divergence between two bivariate normal distributions;\\
\noindent \ding{174} we compute
the left-sided Kullback-Leibler, right-sided Kullback-Leibler, Bhattacharyya
and Fisher-Rao centroids~\cite{nielsen2009sided,BR-2011} (Riemannian geodesic
midpoint); and\\
\noindent \ding{175} we visualize those bivariate normal centroids
(Figure~\ref{fig:mvncentroids}).

The basic usage of \pyBregMan consists of defining a specific Bregman manifold,
defining points on said manifold, and then defining additional geometric
objects on the manifold. 

To install the library via {\tt pip3}, execute the command:
\begin{verbatim}
pip3 install pyBregMan
\end{verbatim}

Alternatively, the library can be installed from {\tt GitHub} as follows:
\begin{verbatim}
git clone https://github.com/alexandersoen/pyBregMan.git
cd pyBregMan
pip3 install .
\end{verbatim}

In the following example, we first define a the
manifold of bivariate normal distributions.
The following code illustrates the construction of a bivariate Gaussian Bregman manifold (see \S\ref{sec:bmtheory} for a quick review of the theory of Bregman manifolds).

\begin{lstlisting}[language=Python, tabsize=4, gobble=8, linerange={4-18}]
import numpy as np

from bregman.application.distribution.exponential_family.gaussian import GaussianManifold
from bregman.base import LAMBDA_COORDS, DualCoords, Point

# Define Bivariate Normal Manifold
manifold = GaussianManifold(input_dimension=2)

# Define data
to_vector = lambda mu, sigma: np.concatenate([mu, sigma.flatten()])
mu_1, sigma_1 = np.array([0.0, 1.0]), np.array([[1.0, 0.5], [0.5, 2.0]])
mu_2, sigma_2 = np.array([1.0, 2.0]), np.array([[2.0, 1.0], [1.0, 1.0]])

point_1 = Point(LAMBDA_COORDS, to_vector(mu_1, sigma_1))
point_2 = Point(LAMBDA_COORDS, to_vector(mu_2, sigma_2))
\end{lstlisting}

Here the manifold bivariate normal distributions is defined via
\VerbLB+GaussianManifold+. In \pyBregMan, we provide a variety of different
manifolds used in information sciences. These can be found under the
\VerbLB+bregman.application+ submodule. To define data on the manifold, one needs
to use the \VerbLB+Point+ primitive. Here data is annotated with a coordinate
type \VerbLB+LAMBDA_COORDS+ (see~\cite{EF-2009}), which for Gaussian distributions corresponds to the
typical mean $\mu$ and covariance matrix $\Sigma$ of Gaussians represented as a
vector.

Simple calculations for (exponential family) distributions can be directly
called from the manifold. For instance, the Kullback-Leibler divergence between
the points can be called as follows:

\begin{lstlisting}[language=Python, tabsize=4, gobble=8, linerange={20-28}, firstnumber=17]
# KL divergence can be calculated
kl = manifold.kl_divergence(point_1, point_2)
rkl = manifold.kl_divergence(point_2, point_1)

print("KL(point_1 || point_2):", kl)
print("KL(point_2 || point_1):", rkl)

# >>> KL(point_1 || point_2): 0.9940936082534253
# >>> KL(point_2 || point_1): 1.2201921060322891
\end{lstlisting}

Additional geometric objects and calculations can be called by external
submodules. For instance, various barycenter (weighted centroid) calculations can be
found in \VerbLB+bregman.barycenter+. Additionally, manifold specific geometric
objects can be found in their specific submodule, see for instance
\VerbLB+FisherRaoKobayashiGeodesic+. Here we are utilizing the fact that the
Fisher-Rao centroid between two points is the midpoint of the corresponding
Fisher-Rao geodesic~\cite{Kobayashi-2023,FR-2023}.

\begin{lstlisting}[language=Python, tabsize=4, gobble=8, linerange={30-60}, firstnumber=26]
from bregman.barycenter.bregman import BregmanBarycenter, SkewBurbeaRaoBarycenter
from bregman.application.distribution.exponential_family.gaussian.geodesic import FisherRaoKobayashiGeodesic

# We can define and calculate centroids
theta_barycenter = BregmanBarycenter(manifold, DualCoords.THETA)
eta_barycenter = BregmanBarycenter(manifold, DualCoords.ETA)
br_barycenter = SkewBurbeaRaoBarycenter(manifold)
dbr_barycenter = SkewBurbeaRaoBarycenter(manifold, DualCoords.ETA)

theta_centroid = theta_barycenter([point_1, point_2])
eta_centroid = eta_barycenter([point_1, point_2])
br_centroid = br_barycenter([point_1, point_2])
dbr_centroid = dbr_barycenter([point_1, point_2])

# Mid point of Fisher-Rao Geodesic is its corresponding centroid of two points
fr_geodesic = FisherRaoKobayashiGeodesic(manifold, point_1, point_2)
fr_centroid = fr_geodesic(t=0.5)

print("Right-Sided Centroid:", manifold.convert_to_display(theta_centroid))
print("Left-Sided Centroid:", manifold.convert_to_display(eta_centroid))
print("Bhattacharyya Centroid:", manifold.convert_to_display(br_centroid))
print("Fisher-Rao Centroid:", manifold.convert_to_display(fr_centroid))

# >>> Right-Sided Centroid: $\mu$ = [0.33333333 1.55555556]; $\Sigma$ = [[1.33333333 0.66666667]
# >>>  [0.66666667 1.11111111]]
# >>> Left-Sided Centroid: $\mu$ = [0.5 1.5]; $\Sigma$ = [[1.75 1.  ]
# >>>  [1.   1.75]]
# >>> Bhattacharyya Centroid: $\mu$ = [0.41772374 1.53238683]; $\Sigma$ = [[1.53973074 0.82458984]
# >>>  [0.82458984 1.40126629]]
# >>> Fisher-Rao Centroid: $\mu$ = [0.5326167  1.62759115]; $\Sigma$ = [[1.72908532 0.98542147]
# >>>  [0.98542147 1.54545598]]
\end{lstlisting}

Finally, we can visualize the objects on the manifold in \VerbLB+matplotlib+
\cite{matplotlib} via using the `MatplotlibVisualizer` class and specifying the
objects to plot. One can also register additional callback functions to enhance
the visualization. Here we utilize a Tissot-style visualization of the
covariance matrices whilst plotting the mean vector. The final visualization is
shown in Figure~\ref{fig:mvncentroids}. Notice that the \VerbLB+plot_object+
function accepts typical \VerbLB+matplotlib+ arguments.

\begin{lstlisting}[language=Python, tabsize=4, gobble=8, linerange={62-79}, firstnumber=57]
from bregman.visualizer.matplotlib.callback import VisualizeGaussian2DCovariancePoints
from bregman.visualizer.matplotlib.matplotlib import MatplotlibVisualizer

# These objects can be visualized through matplotlib
visualizer = MatplotlibVisualizer(manifold, (0, 1))
visualizer.plot_object(point_1, c="black")
visualizer.plot_object(point_2, c="black")
visualizer.plot_object(
    theta_centroid, c="red", label="Right-Sided Centroid"
)
visualizer.plot_object(eta_centroid, c="blue", label="Left-Sided Centroid")
visualizer.plot_object(
    br_centroid, c="purple", label="Bhattacharyya Centroid"
)
visualizer.plot_object(fr_centroid, c="pink", label="Fisher-Rao Centroid")
visualizer.add_callback(VisualizeGaussian2DCovariancePoints())

visualizer.visualize(LAMBDA_COORDS)  # Display coordinate type
\end{lstlisting}

\section{A short introduction to Bregman and Fisher-Rao manifolds }\label{sec:bmtheory}

We first briefly review the concepts of Bregman manifolds and its relationship with  Fisher-Rao manifolds in \S\ref{sec:BM}.
Then we explain the goals of the \pyBregMan library in \S\ref{sec:goals}.
Let us mention the {\tt GeomStats} Python library which includes an information geometry package~\cite{IGGeomstats-2023} implementing Fisher-Rao manifolds, and the {\tt Manifolds.jl} Julia package~\cite{axen2023manifolds} among other public libraries.

\subsection{Bregman manifolds: A quick review}\label{sec:BM}

Let $F:\Theta\subset\rightarrow\bbR$ be a strictly convex and differentiable real-valued convex function defined over a topologically open domain $\Theta\subset\bbR^m$.
The Bregman divergence~\cite{Bregman-1967} between $\theta_1\in\Theta$ and $\theta_2\in\Theta$ is a measure of dissimilarity defined by the following formula:
$$
B_F(\theta_1:\theta_2)=F(\theta_1)-F(\theta_2)-\inner{\theta_1-\theta_2}{\nabla F(\theta_2)},
$$
where $\inner{x}{y}=x^\top y=\sum_{i=1}^m x_iy_i$ is the Euclidean inner product called dot product or scalar product.

When the Bregman generator $F$ inducing the Bregman divergence $B_F$ is $C^3$ and of Legendre-type~\cite{Rockafellar-1967}, a Bregman divergence induces a dually flat space $(M,g,\nabla,\nabla^*)$ in information geometry~\cite{IG-2016} with a single global chart $(\Theta,\theta(\cdot))$.

The torsion-free affine connections $\nabla$ and $\nabla^*$ induced respectively by $F(\theta)$ and its convex conjugate\footnote{The convex conjugate $F^*(\eta)=\sup_{\theta\in\Theta} \inner{\theta}{\eta}-F(\theta)$ of a Legendre-type function $(\Theta,F(\theta))$ is of Legendre-type, and the Legendre-Fenchel transform is involutive. We have two gradient maps: $\eta(\theta)=\nabla F(\theta)$ and $\theta(\eta)=\nabla F^*(\eta)$ which are reciprocal to each other (i.e., $\nabla F=(\nabla F)^{-1}$), and the convex conjugate is given by the Legendre function
$F^*(\eta)=\inner{\nabla F^{-1}(\eta)}{\eta}-F(\nabla F^{-1}(\eta))$.}  
$F^*(\eta)$ are flat and coupled to the metric tensor $g$.
The metric $g$ is of type Hessian and can be expressed either in the $\theta$-coordinate system or the dual $\eta$-coordinate system equivalently by
$$
g(\theta)=\nabla^2 F(\theta),\quad g(\eta)=\nabla^2 F^*(\eta).
$$
Dually flat spaces are special cases of Hessian manifolds~\cite{Shima-2007} which possibly admit several charts and local dual potential functions $F$ and $F^*$ related by the Legendre-Fenchel transformation.

We term Hessian manifolds with single charts or dually flat spaces {\em Bregman manifolds} because conversely, a single chart manifold $M$ with a dual structure $(M,g,\nabla,\nabla^*)$ admits a pair of canonical potential functions $F(\theta)$ and $F^*(\eta)$ such that its coordinate-free dually flat divergence $D_{\nabla,\nabla^*}$ between points $p_1$ and $p_2$ of $M$ (with $\theta_i=\theta(p_i)$ and $\eta_i=\eta(p_i)$) can be expressed equivalently as:
$$
D_{\nabla,\nabla^*}(p_1:p_2)=B_F(\theta_1:\theta_2)=B_{F^*}(\eta_2:\eta_1)=Y_{F,F^*}(\theta_1,\eta_2)=Y_{F^*,F}(\eta_2,\theta_1),
$$
where $Y_{F,F^*}(\theta_1,\eta_2)$ is the Fenchel-Young divergence~\cite{FY-2020,FY-2022}:
$$
Y_{F,F^*}(\theta_1,\eta_2)=F(\theta_1)+F^*(\eta_2)-\inner{\theta_1}{\eta_2}.
$$
See~\cite{ManyIG-2022} for more details on Bregman manifolds.

Notice that the potential functions $F(\theta)$ and $F^*(\eta)=:G(\eta)$ reconstructed from
$(M,g,\nabla,\nabla^*)$ are not unique~\cite{IG-2016} but defined up to affine terms. That is, we may consider equivalently the primal potential function:
$$
\barF(\theta)=F(A\theta+b)+ \inner{c}{\theta}+d
$$
for $A\in\GL(d,\bbR)$, $b, c\in\bbR^m$ and  $d\in\bbR$, and its corresponding convex conjugate ${\bar{F}}^*$ to build the same information-geometric structure $(M,g,\nabla,\nabla^*)$. 
This property has been called affine Legendre equivalence in~\cite{singularDFS-2021}. See \S{}4 of~\cite{QAC-2023} for more details.

Bregman manifolds can be constructed from convex functions arising from statistical models.
For example,
\begin{itemize}

\item When $F(\theta)$ is the cumulant function of an exponential family~\cite{IG-2016}, we get a Bregman manifold with Bregman divergence corresponding to the reverse Kullback-Leibler divergence~\cite{Nielsen-2024}.
When choosing the partition function $Z(\theta)=\exp(F(\theta))$ instead of $F(\theta)$ (a log-convex function hence convex), the Bregman divergence amounts to the Kullback-Leibler divergence between unnormalized densities~\cite{ZF-2024}.

\item When $F(\theta)$ is the negentropy of a mixture family~\cite{IG-2016}, we get a Bregman manifold with Bregman divergence corresponding to the Kullback-Leibler divergence~\cite{MCIG-2019}.

\item All statistical models parameterized by either one or two parameters can yield Bregman manifolds because the 2D Fisher-Rao metrics are Hessian. See for example the case of univariate elliptical distributions~\cite{Mitchell-1988}.
\end{itemize}

\subsection{Fisher-Rao manifolds}

A Fisher-Rao manifold~\cite{Rao-1945} $(M,g)$ is a Riemannian manifold $M$ induced by the Fisher information metric $g$ of a statistical model 
$\{p_\theta(x) \st\theta\in\Theta\}$ which yields a geodesic distance called Rao's distance~\cite{BurbeaRao-1982}.
When the Fisher information metric is Hessian and defined according to a single chart (i.e., there exists a coordinate system $\theta(\cdot)$ and a potential function $F(\theta)$ such that $g(\theta)=\nabla^2 F(\theta)$), the Fisher-Rao manifold can also be considered as a Bregman manifold.

\begin{figure}
    \centering
    \includegraphics[width=0.5\textwidth]{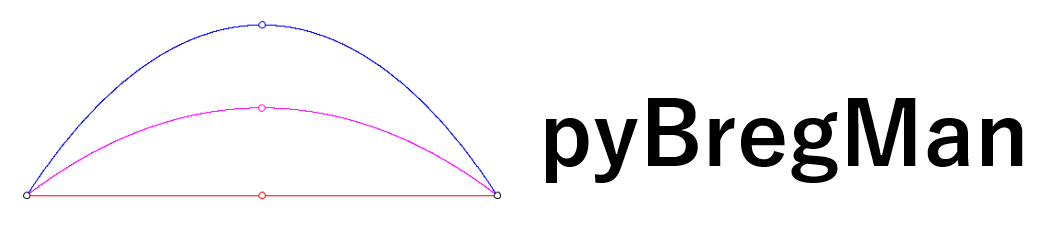}
    \caption{The logo for the \pyBregMan library displays the Fisher-Rao geodesics (self-dual $0$-geometry) and the dual exponential/mixture geodesics 
 ($\pm 1$-geometry) between two univariate normal distributions visualized on the $(\mu,\sigma)$ upper plane.}\label{fig:logo}
\end{figure}

For example, the Fisher-Rao manifold of the statistical model of univariate Gaussian distributions is Hessian since parameterized by a 2D parameter.
This manifolds is also  a Bregman manifold.
Figure~\ref{fig:logo} for illustrates the Fisher-Rao geodesic (purple), $\nabla$-geodesic (called $e$-geodesic in blue induced by the cumulant function $F$) and the $\nabla^*$-geodesic (called $m$-geodesic in red) used for the logo of \pyBregMan. The $e/m$-geodesics correspond to the $\pm 1$-geometry and the Fisher-Rao geodesic to the self-dual $0$-geometry, see~\cite{IG-2016}.

\subsection{Divergence-based information geometry}

Given a smooth divergence $D(\theta_1:\theta_2)$, Eguchi~\cite{Eguchi-1992} proposed a method to induce a divergence-based manifold, i.e., a dualistic structure $(M,g,\nabla,\nabla^*)$ related to the forward divergence $D(\theta_1:\theta_2)$ and reverse divergence $D^*(\theta_1:\theta_2)=D(\theta_2:\theta_1)$. 
In particular, when $D(\theta_1:\theta_2)=B_F(\theta_1:\theta_2)$ is a Bregman divergence, the divergence-based manifold amounts to a Bregman manifold.

\begin{figure}
    \centering
    \includegraphics[width=\textwidth]{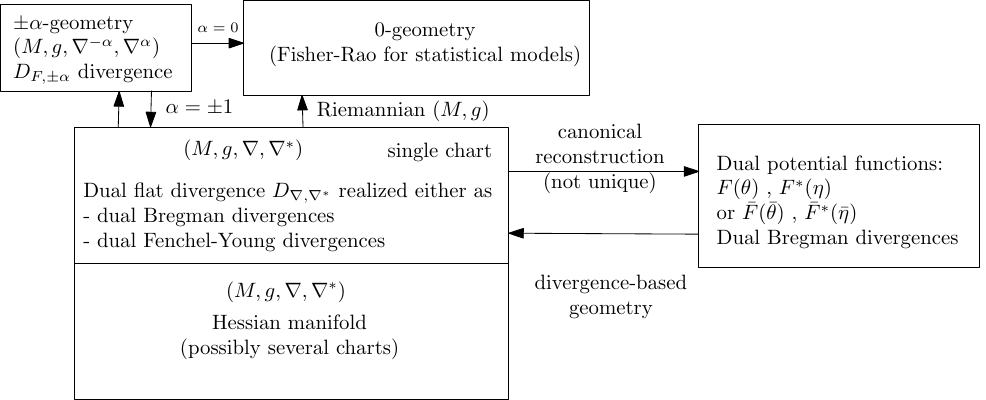}
    \caption{Overview of the structures with related divergences.}\label{fig:structures}
\end{figure}

When $D(\theta_1:\theta_2)=D_{F,\alpha}(\theta_1:\theta_2)$ is Zhang $\alpha$-divergence~\cite{Zhang-2004} (also called Burbea-Rao divergence in~\cite{BR-2011}):
$$
D_{F,\alpha}(\theta_1:\theta_2)=\frac{4}{1-\alpha^2}\left(
    \frac{1-\alpha}{2}F(\theta_1)+\frac{1+\alpha}{2}F(\theta_2)-
    F\left(\frac{1-\alpha}{2}\theta_1+\frac{1+\alpha}{2}\theta_2\right)
\right),\quad \alpha\in\bbR\backslash\{-1,1\},
$$
the divergence-based manifold induced by $D_{F,\alpha}(\theta_1:\theta_2)$ is a $\alpha$-Hessian manifold~\cite{Zhang-2014} which corresponds to the $\alpha$-geometry~\cite{IG-2016} extending the dual Bregman manifolds which are obtained in the limit cases $\alpha=1$ and $\alpha=-1$:
$$
B_{F}(\theta_1:\theta_2)=\lim_{\alpha\rightarrow -1} D_{F,\alpha}(\theta_1:\theta_2),
\quad
B_{F}(\theta_2:\theta_1)=\lim_{\alpha\rightarrow 1} D_{F,\alpha}(\theta_1:\theta_2).
$$
By continuity, we let $D_{F,-1}(\theta_1:\theta_2)=B_{F}(\theta_1:\theta_2)$ 
and  $D_{F,1}(\theta_1:\theta_2)=B_{F}(\theta_2:\theta_1)$.
Figure~\ref{fig:structures} shows how structures and divergences relate to each others.

The $0$-Hessian manifold correspond to the Fisher-Rao manifold when the Fisher metric is Hessian, and the statistical divergence corresponding to $D_{F,0}$ induced by the cumulant function of an exponential family is the Bhattacharyya distance~\cite{BR-2011}.

\subsection{Purposes of the library}\label{sec:goals}

The three main goals of the \pyBregMan library are

\begin{enumerate}
    \item To provide a {\em geometric kernel} to implement, visualize, and export figures of main concepts of Bregman manifolds in 2D and 3D (e.g., dual $\theta/\eta$ coordinates which are $\nabla/\nabla^*$-affine coordinate systems, reciprocal basis, connection-based geodesics, metric-compatible parallel transport, Bregman balls, Bregman projections, etc.);

    \item To provide APIs for handling the Fisher-Rao manifolds and the dualistic structure ($\alpha$-geometry) in information geometry of statistical models with Fisher information of Hessian type;

    \item To use the APIs to demonstrate use cases of information geometry on exponential and mixture families: 
        For example, to compute the Chernoff information~\cite{Chernoff-2022}, 
        to approximate the matrix geometric mean~\cite{nakamura2001algorithms}, or to compute the discrete Jensen-Shannon centroid~\cite{JSCentroid-2020}, etc.
        That is, to ease the programming of information geometry in action in various fields of information sciences.
\end{enumerate}

In particular, we shall consider the following Bregman manifolds:

\begin{itemize}
    \item The family categorical distributions interpreted either as an
        exponential family or a (dual) mixture family~\cite{IG-2016};

    \item The exponential family of multivariate normal distributions for which
        the Fisher-Rao geodesics~\cite{Kobayashi-2023} have been recently
        elicited, yielding efficient $(1+\epsilon)$-approximation algorithms for the
        Fisher-Rao distance~\cite{FisherRao-Nielsen-2024} for any $\epsilon>0$;

    \item The characteristic functions of regular cones including the matrix
        symmetric positive definite (SPD) cone which find applications in
        mathematical programming~\cite{ohara2024doubly} among others.
\end{itemize}

\section{Structure of the library}

In the following section, we outline the structure of the \pyBregMan library.
Additionally, we highlight connections between specific implementation details
and API design which corresponds to the concepts of Bregman manifolds, \ie,
Section~\ref{sec:BM}. We first highlight core primitives of the library,
clarifying how manifold objects and data interact within \pyBregMan.
We then outline additional geometric objects which are currently included in the library.
Defined application manifolds which can be readily used are presented. Finally,
we present the visualization pipeline.

\subsection{Primitives of the Library}

Various primitives of \pyBregMan are found in the \VerbLB+bregman.base+ and
\VerbLB+bregman.manifold+ submodules. We will examine the \VerbLB+Coords+ type and
the \VerbLB+BregmanManifold+ abstract class.

\subsubsection{Coordinates and Points}
From the former, an important aspect of the library is how the \VerbLB+Coords+
class annotates different geometric objects, which inherit from
\VerbLB+CoordObject+. For instance, \VerbLB+Point+ provides a coordinate type
annotation for data to be used in our manifolds. All Bregman manifolds are
equipped with \VerbLB+THETA_COORDS+ ($\theta$-coordinate system) and
\VerbLB+ETA_COORDS+ ($\eta$-coordinate system) which are instances of coordinates
\VerbLB+Coords+. Additional instances of \VerbLB+Coords+ can be defined, for
instance \VerbLB+LAMBDA_COORDS+ (ordinary parameters) for Gaussian distributions,
see Section~\ref{sec:getting-started}. A subset of coordinates are defined by
\VerbLB+DualCoords+, which provide an enum type restricted to only $\theta$-and
$\eta$-coordinate systems. This provides restriction of the \pyBregMan API for
geometric objects which are only defined \wrt $\theta$-and
$\eta$-coordinate systems.

Notice that in the discussion of the coordinate system typing of the library,
no specific manifold has been specified. Indeed, currently data defined by
\VerbLB+Point+ can be freely switched between different manifolds under the
assumption that the dimension (shape of the data) and constraints (\eg,
ensuring positive semi-definiteness of Gaussian covariances) of the data is
valid in each manifold.

For example, the dually flat symmetric positive-definite manifold~\cite{IG-2016} can be obtained either as the exponential family manifold of same-mean Gaussian distributions or as the Bregman manifold induced by the logdet barrier function~\cite{ohara1996dualistic}.

\subsubsection{Bregman Manifolds}
\begin{figure}
    \centering
    \includegraphics[width=1.0\textwidth]{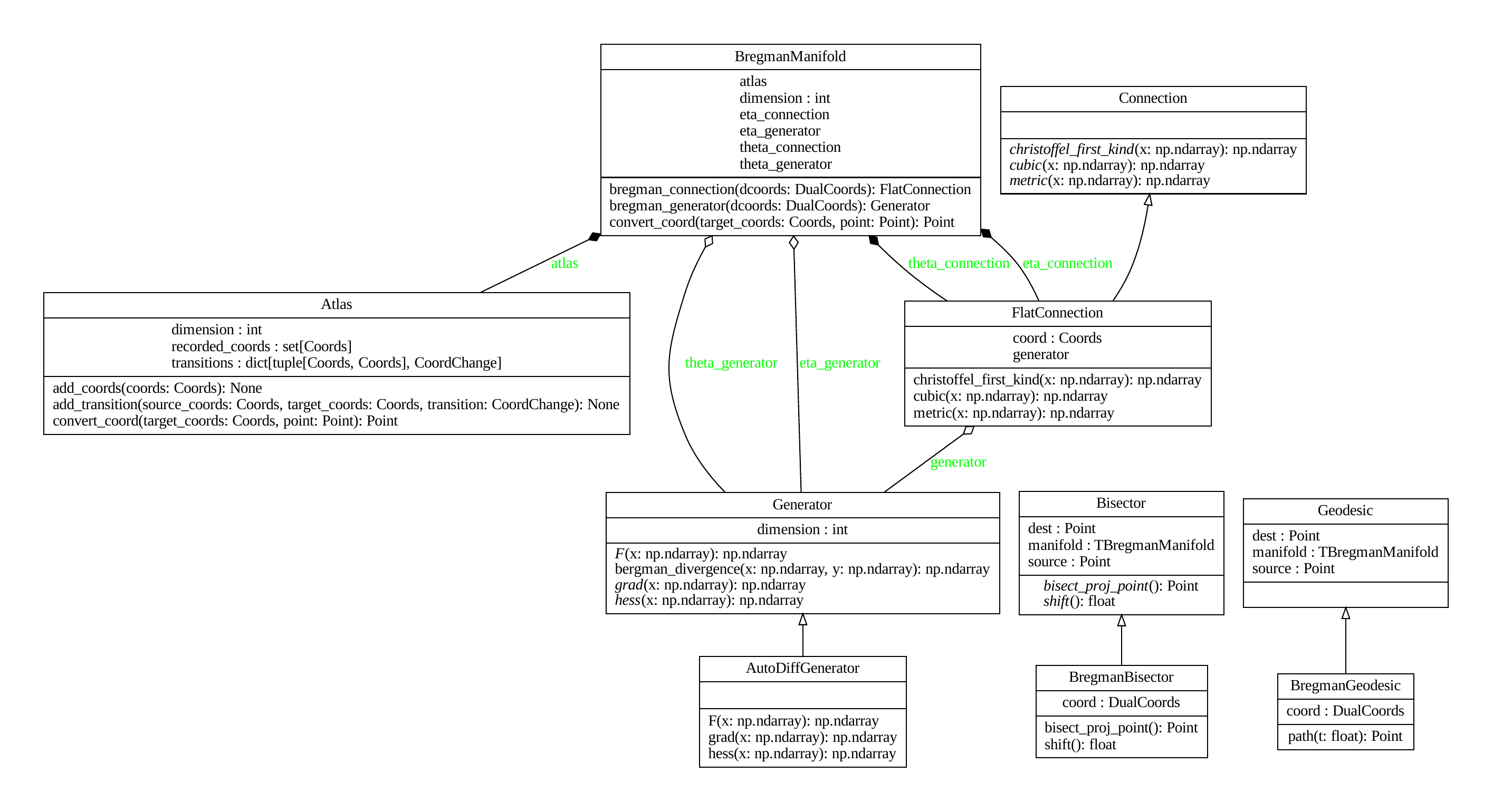}
    \caption{%
        UML dependency graph of the {\tt bregman.manifold.manifold} submodule. Zoom in for details.
    }\label{fig:class-BregmanManifold}
\end{figure}

The \VerbLB+bregman.manifold+ submodule provides the main components of all
manifold classes defined in the library. In particular, all manifold objects in
\pyBregMan implement from the abstract class \VerbLB+BregmanManifold+ in
\VerbLB+bregman.manifold.manifold+. The dependency graph of
\VerbLB+BregmanManifold+ is shown in Figure~\ref{fig:class-BregmanManifold}.
A \VerbLB+BregmanManifold+ must define two Bregman \VerbLB+Generator+s and define
the \VerbLB+dimension+ of the generators' inputs. The two generators consist of a
\VerbLB+theta_generator+ $F$ and a \VerbLB+eta_generator+ $F^*$. \pyBregMan assumes
that the defined generator classes are dual. An \VerbLB+eta_generator+ is
optional in its definition and can be set to \VerbLB+None+; however, this would
greatly limit the functionality of the manifold.

The prescribed \VerbLB+Generator+ class requires $F$/$F^*$ to be twice
differentiable, with their corresponding derivatives defined, see its
corresponding methods in Figure~\ref{fig:class-BregmanManifold}. 
For quick development and easier usage, we provide \VerbLB+AutoDiffGenerator+ in
\VerbLB+bregman.manifold.generator+, which uses the \VerbLB+autograd+ library to
automatically calculate the required first and second derivatives via
auto-differentiation. As such, by using \VerbLB+AutoDiffGenerator+, a user only
needs to define $F$ with gradient $\nabla F$ and Hessian $\nabla^2 F$ automatically defined.
We do note that for numeric precision, one may still want to define the
derivatives explicitly.

The \VerbLB+FlatConnection+ composition in \VerbLB+BregmanManifold+ simply defines
the torsion-free affine connections induced by the input \VerbLB+Generator+s
(utilizes auto-differentiation to calculate the metric).

Finally, the \VerbLB+Atlas+ composition in \VerbLB+BregmanManifold+ is a class
which provides management of the different coordinate systems specified by
\VerbLB+Coords+. In particular, it allows instances of \VerbLB+Coords+ and
coordinate changes to be registered in the \VerbLB+BregmanManifold+. This allows
coordinate dependent geometric operations to be freely utilized on data which
may potentially have different coordinates. We show instances of this
convenience in Section \ref{subsec:geometric-objects}.
It should be noted to allow coordinate changes between all registered
coordinate systems, all combinations of coordinate changes must be defined.
The conversion between $\theta$-and $\eta$-coordinates are automatically
defined via $\nabla F$ and $\nabla F^*=\nabla G$, respectively.

\subsubsection{Geodesics and Bisectors}

Two additional geometric objects that are defined in the
\VerbLB+bregman.manifold+ submodule are the \VerbLB+Geodesic+ and \VerbLB+Bisector+
objects~\cite{BVD-2010}. For the latter, \VerbLB+Bisector+ is currently only for visualization,
see Section~\ref{subsec:chernoff-example}.
\VerbLB+Geodesic+ defines an abstract object, which when implemented provides a
curve object which returns points when given a value $t \in [0, 1]$, mapping
from source ($t = 0$) to destination ($t = 1$). In
\VerbLB+bregman.manifold.geodesic+, a concrete implementation of
\VerbLB+BregmanGeodesic+ is defined. 
This allows a $\theta$-or $\eta$-geodesic to be defined for a specified
manifold. 
Its implementation is shown in the following.

\begin{lstlisting}
from bregman.base import Curve, DualCoords, Point
from bregman.manifold.manifold import BregmanManifold

...

class BregmanGeodesic(Geodesic[BregmanManifold]):

    def __init__(self, manifold: BregmanManifold, source: Point, dest: Point, dcoords: DualCoords = DualCoords.THETA) -> None:
        super().__init__(manifold, source, dest)

        self.coord = dcoords

    def path(self, t: float) -> Point:
        src_coord_data = self.manifold.convert_coord(self.coord.value, self.source).data
        dst_coord_data = self.manifold.convert_coord(self.coord.value, self.dest).data

        # As flat in its own coordinate
        return Point(
            coords=self.coord.value,
            data=(1 - t) * src_coord_data + t * dst_coord_data,
        )
\end{lstlisting}

In addition the \VerbLB+source+ and \VerbLB+dest+ data \VerbLB+Point+s, it takes in a
\VerbLB+BregmanManifold+ and a \VerbLB+DualCoord+. This allows the source and
destination point of the geodesic to be converted into the specified dually
flat coordinate (as per \VerbLB+dcoords: DualCoords+) to define the geodesic via
linear interpolation.

\subsection{Additional Geometric Objects}
\label{subsec:geometric-objects}

\begin{figure}
    \centering
    \includegraphics[width=1.0\textwidth]{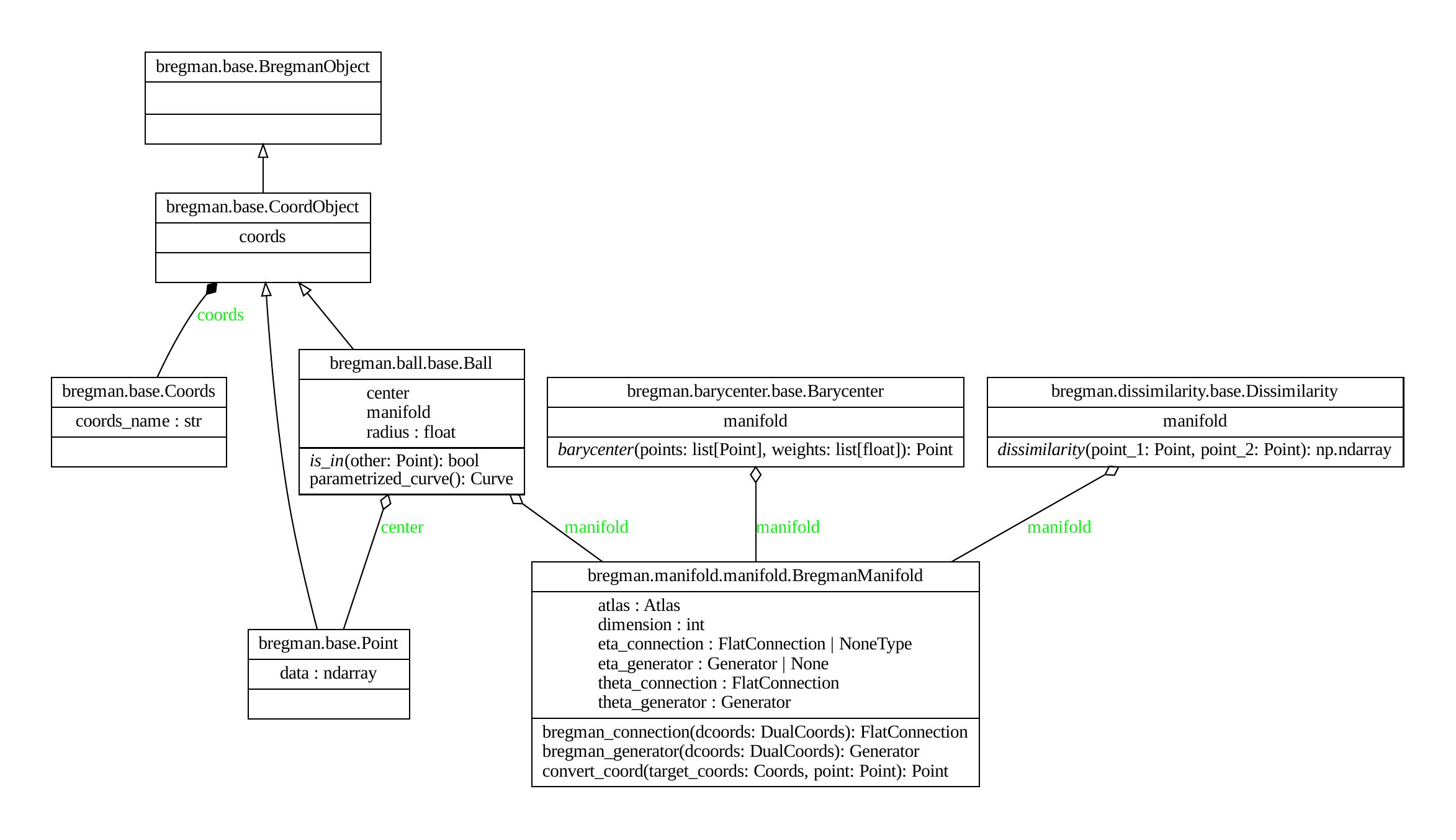}
    \caption{%
        Abridged UML dependency graph of additional geometric objects.
    }\label{fig:application}
\end{figure}

In the following, we provide a brief description the additional geometric
objects provided by \pyBregMan. We   only discuss their abstract classes and
leave example usages to Section~\ref{sec:examples}. The geometric objects can
be depicted in Figure~\ref{fig:application}.

\begin{figure}
    \centering
    \includegraphics[width=0.7\textwidth]{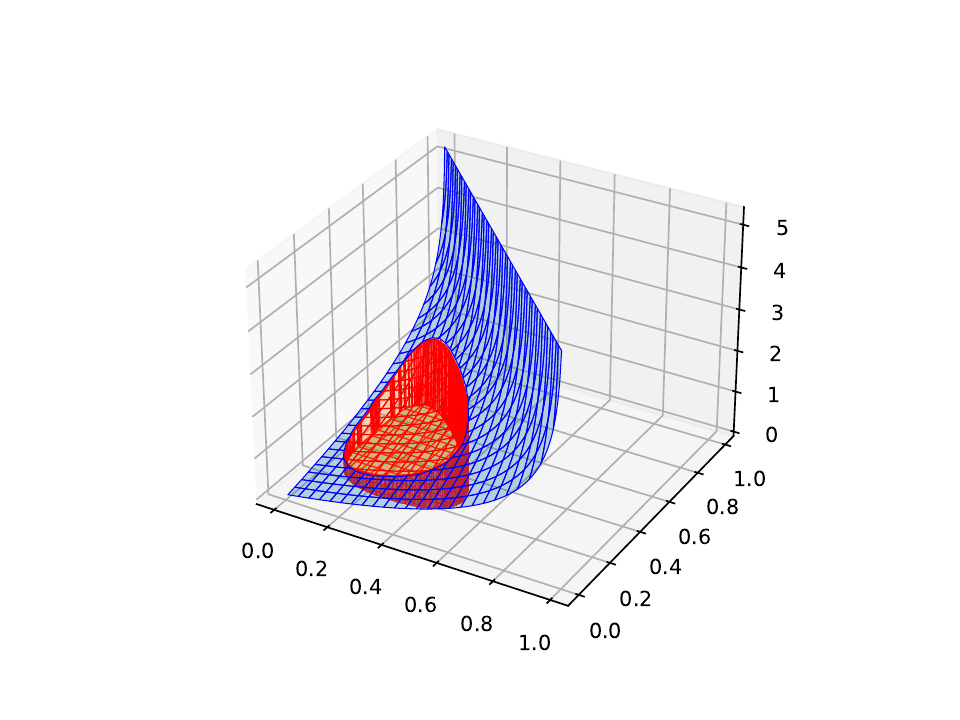}
    \caption{%
        A 2D Bregman ball/sphere can be obtained as the vertical projection of the intersection of a 3D plane with the graph of the potential function~\cite{BVD-2010}.
    }\label{fig:bball}
\end{figure}

\begin{itemize}
    \item \VerbLB+bregman.ball.base.Ball+: The \VerbLB+Ball+ class allows objects
        such as Bregman balls to be defined in the library. It is defined from
        a ball \VerbLB+center+ and a \VerbLB+radius+. Currently, the functionality
        of the class is limited to checking whether or not a point is inside
        the ball via the \VerbLB+is_in+ method. Unlike the other geometric
        objects discussed in this subsection, \VerbLB+Ball+ is a
        \VerbLB+CoordObject+. This means that \VerbLB+Ball+ must have a ``natural''
        coordinate system attributed to its. 
Bregman balls are basic geometric objects used in proximity query data structures~\cite{BregmanVPT-2009,BBtree-2009}. The smallest enclosing Bregmall ball of a set of points has been studied in~\cite{nielsen2008smallest,SEBB-2005}.
Bregman balls can be considered as intersections of hyperplanes with the 
 graph of the potential function~\cite{BVD-2010}, see Figure~\ref{fig:bball}.
        
    \item \VerbLB+bregman.barycenter.base.Barycenter+: The \VerbLB+Barycenter+
        class provides an interface for different barycenter algorithms to be
        defined. In particular, for a particular manifold, it implements a
        method \VerbLB+barycenter+ which ``aggregates'' a list of points
        \VerbLB+list[Point]+ and a list of weights \VerbLB+list[float]+ (of the
        same length). Examples of concrete implementations of \VerbLB+Barycenter+
        can be found in Section~\ref{sec:getting-started}.
    \item \VerbLB+bregman.dissimilarity.base.Dissimilarity+: The
        \VerbLB+Dissimilarity+ class provides an interface for the definition of
        dissimilarity functions in the library. These are the class of function
        which accepts two points and calculates a dissimilarity measure. An
        example of a concrete implementation of \VerbLB+Dissimilarity+ is the
        \VerbLB+BregmanDivergence+.
        In information geometry, a smooth dissimilarity is called a divergence and interpreted as a function on the product manifold: a yoke~\cite{Yoke-1997} or contrast function~\cite{Matumoto-1993}.
\end{itemize}

\subsection{Application Manifolds}

We provide a variety of application manifolds in \pyBregMan. These manifolds
are concrete implementations of the \VerbLB+BregmanManifold+. In particular, they
all inherit from the \VerbLB+ApplicationManifold+ class in the submodule
\VerbLB+bregman.application.application+. Implementing \VerbLB+ApplicationManifold+
has two additional requirements over \VerbLB+BregmanManifold+. First, a
$\lambda$-coordinate system needs to be defined. This is typically a representation of
points in the manifold which do not correspond to either the $\theta$-or
$\eta$-coordinates. Secondly, a \VerbLB+DisplayPoint+ needs to be defined. This
is just a pretty-printing wrapper for a specific coordinate system.

The following presents a list of implemented application manifolds.

\begin{itemize}

    \item Distributions (statistical models):
        \begin{itemize}
            \item Exponential Family
                \begin{itemize}
                    \item \VerbLB+bregman.application.distribution.exponential_family+\\
                          \VerbLB+       .gaussian.GaussianManifold+
                    \item \VerbLB+bregman.application.distribution.exponential_family+\\
                          \VerbLB+       .multinomial.MultinomialManifold+
                    \item \VerbLB+bregman.application.distribution.exponential_family+\\
                          \VerbLB+       .categorical.CategoricalManifold+
                \end{itemize}
            \item Mixture 
                \begin{itemize}
                    \item \VerbLB+bregman.application.distribution.mixture+\\
                          \VerbLB+       .discrete_mixture.DiscreteMixture+
                    \item \VerbLB+bregman.application.distribution.mixture+\\
                          \VerbLB+       .ef_mixture.EFMixtureManifold+
                \end{itemize}
        \end{itemize}
    \item \VerbLB+bregman.application.psd.PSDManifold+
\end{itemize}

In each submodule, additional subclasses of \VerbLB+ApplicationManifold+ are
defined and used. For instance, all \VerbLB+DistributionManifold+s require a
mapping from \VerbLB+Point+ to \VerbLB+Distribution+ (from
\VerbLB+bregman.object.distribution+).

In addition to the manifold defined, additional manifold specific geometric
objects are sometimes defined. For instance, the \VerbLB+GaussianManifold+
contains additional submodules which defines Gaussian specific dissimilarity
measures and geodesics. Notably, we provide implementations of approximate
Fisher-Rao distance and geodesics using Kobayashi's approximation \cite{Kobayashi-2023}.

\subsection{Visualization}

\pyBregMan provides a visualization submodule in \VerbLB+bregman.visualizer+. In
particular, \VerbLB+MatplotlibVisualizer+ in \VerbLB+bregman.visualizer.matplotlib+
provides an implementation of a visualizer using the \VerbLB+matploblib+
library~\cite{matplotlib}. The visualizer implements a plot function for every
subtype of \VerbLB+BregmanObject+ (defined in \VerbLB+bregman.base+). These are
points \VerbLB+Point+, geodesics \VerbLB+Geodesic+, and bisectors \VerbLB+Bisector+.
Additional \VerbLB+VisualizerCallback+s can be defined if additional plotting
functionality is desired.
Also note that \VerbLB+matplotlib+ specific arguments can also be used when using
\VerbLB+MatplotlibVisualizer+, see Section~\ref{sec:getting-started}.

\section{Some examples of Hessian Fisher-Rao statistical models}
\label{sec:examples}

In the following section, we present a series of visual vignettes to
demonstrate different uses of \pyBregMan. In particular, we consider the following examples:
\begin{itemize}
    \item Positive Semi-definite Matrices (SPD cone): Inductive arithmetic-harmonic means~\cite{Nakamura-2001} to calculate the geometric matrix mean (see also~\cite{inductivemean-2023}). See Figure~\ref{fig:ahm}.
    
    \item Categorical Distributions: Calculating pixel intensity histogram centroids of images. In particular, we calculate the sided Kullback-Leibler centroids~\cite{nielsen2009sided} and the Jensen-Shannon centroid~\cite{JSCentroid-2020}.

    \item Gaussian Distributions: Calculating the Chernoff point~\cite{Chernoff-2022} which is useful in Bayesian hypothesis testing and in information fusion tasks.
\end{itemize}

\subsection{Positive Semi-definite Matrices: Geometric Matrix Means}

In the space of Positive Semi-definite (PSD) matrices, an iterative algorithm
exists to obtain the mid-point of the Riemannian geodesic. In particular, it
has been shown that by iteratively taking the arithmetic and harmonic mean
between PSD matrices, and then setting those means as the new ``initial''
points, one will converge in a quadratic order to the geometric
mean~\citep{Nakamura-2001}.
For PSD matrices, the geometric mean between matrices corresponds to the
mid-point of the Riemannian geodesic; and the arithmetic mean and geometric
mean correspond to the mid points of a primal-dual pair of geodesics.

\begin{figure}
    \centering
    \includegraphics[width=1.0\textwidth]{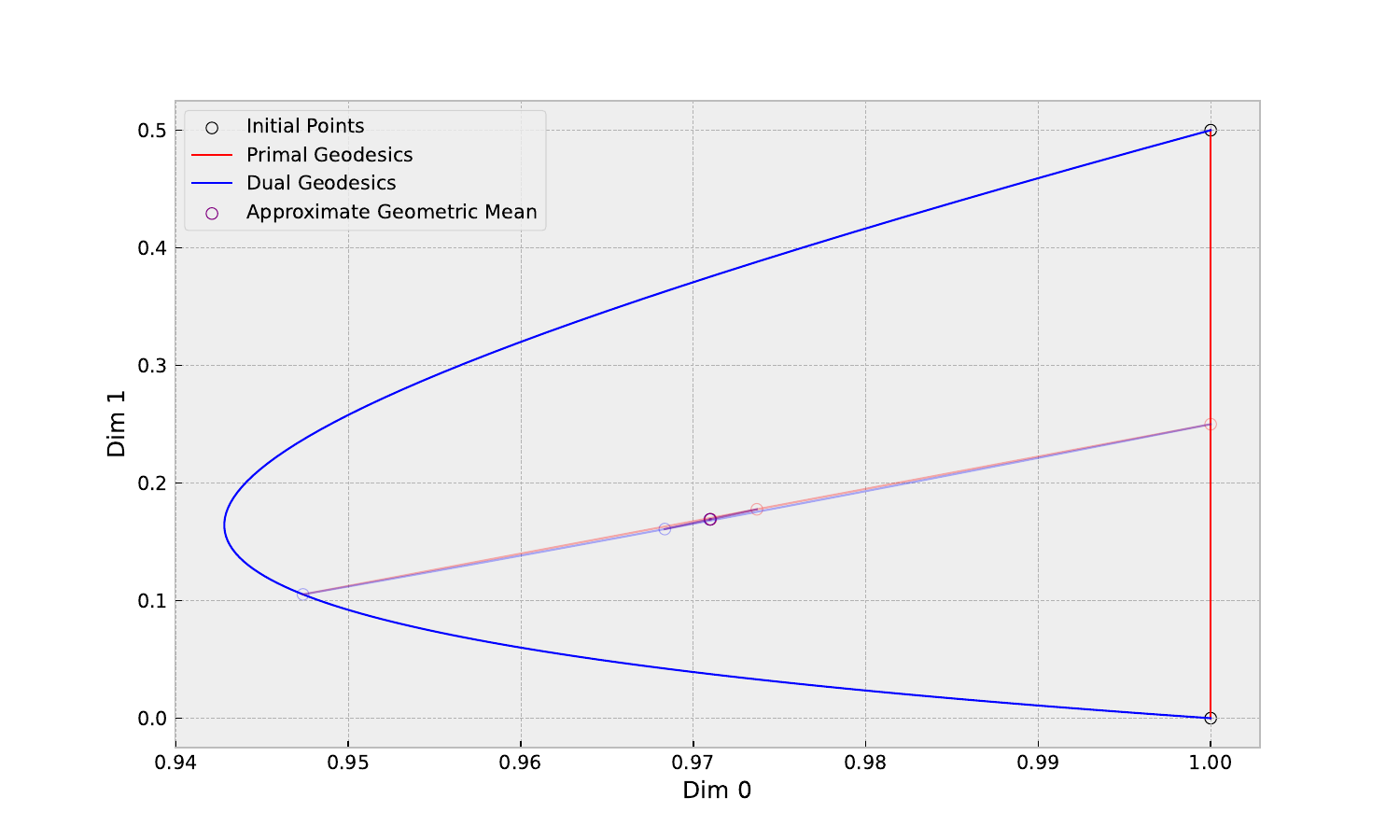}
    \caption{The inductive arithmetic-harmonic mean (AHM) converges to the geometric matrix mean.}
    \label{fig:ahm}
\end{figure}

Figure~\ref{fig:ahm} demonstrates the inductive arithmetic-harmonic mean (AHM)
algorithm for calculating the Riemannian geodesic mid-point. In particular,
note the visualization of the primal and dual geodesic being plotted with their
corresponding mid-points becoming the next starting points.

\begin{lstlisting}
import numpy as np

from bregman.application.psd import PSDManifold
from bregman.base import ETA_COORDS, LAMBDA_COORDS, THETA_COORDS, DualCoords, Point
from bregman.manifold.geodesic import BregmanGeodesic
from bregman.visualizer.matplotlib import MatplotlibVisualizer

DISPLAY_TYPE = LAMBDA_COORDS
VISUALIZE_INDEX = (0, 1)

manifold = PSDManifold(2)

coord1 = Point(LAMBDA_COORDS, np.array([1, 0.5, 2]))
coord2 = Point(LAMBDA_COORDS, np.array([1, 0, 0.5]))

primal_geo = BregmanGeodesic(
    manifold, coord1, coord2, dcoords=DualCoords.THETA
)
dual_geo = BregmanGeodesic(
    manifold, coord1, coord2, dcoords=DualCoords.ETA
)

# Define visualizer
visualizer = MatplotlibVisualizer(manifold, VISUALIZE_INDEX)

# Add objects to visualize
visualizer.plot_object(coord1, c="black", label="Initial Points")
visualizer.plot_object(coord2, c="black")
visualizer.plot_object(primal_geo, c="red", label="Primal Geodesics")
visualizer.plot_object(dual_geo, c="blue", label="Dual Geodesics")

p, q = coord1, coord2
ITERS = 5
for i in range(ITERS):
    primal_geo = BregmanGeodesic(manifold, p, q, dcoords=DualCoords.THETA)
    dual_geo = BregmanGeodesic(manifold, p, q, dcoords=DualCoords.ETA)

    if i > 0:
        visualizer.plot_object(p, c="red", alpha=0.3)
        visualizer.plot_object(q, c="blue", alpha=0.3)

        visualizer.plot_object(primal_geo, c="red", alpha=0.3)
        visualizer.plot_object(dual_geo, c="blue", alpha=0.3)

    p = primal_geo(0.5)
    q = dual_geo(0.5)

visualizer.plot_object(p, c="purple", label="Approximate Geometric Mean")

visualizer.visualize(DISPLAY_TYPE)
\end{lstlisting}

\subsection{Categorical Distributions: Intensity Centroids}

In the following example, we utilize \pyBregMan to generate different types of
centroid on the categorical distribution manifold. In particular, we consider the
Jensen-Shannon centroid and its corresponding dual (Jeffreys Centroid)~\citep{JSCentroid-2020}. As
we are working in the categorical manifold, we will utilize the convenient
connection that the manifold of (discrete) mixture distributions is dual to the
categorical manifold; and that the Jensen-Shannon divergences can be calculated
as specific instances of the Burbea-Rao divergences on this mixture
manifold~\citep{BR-2011}.

\begin{figure}
    \centering
    \includegraphics[width=0.4\textwidth, angle=270]{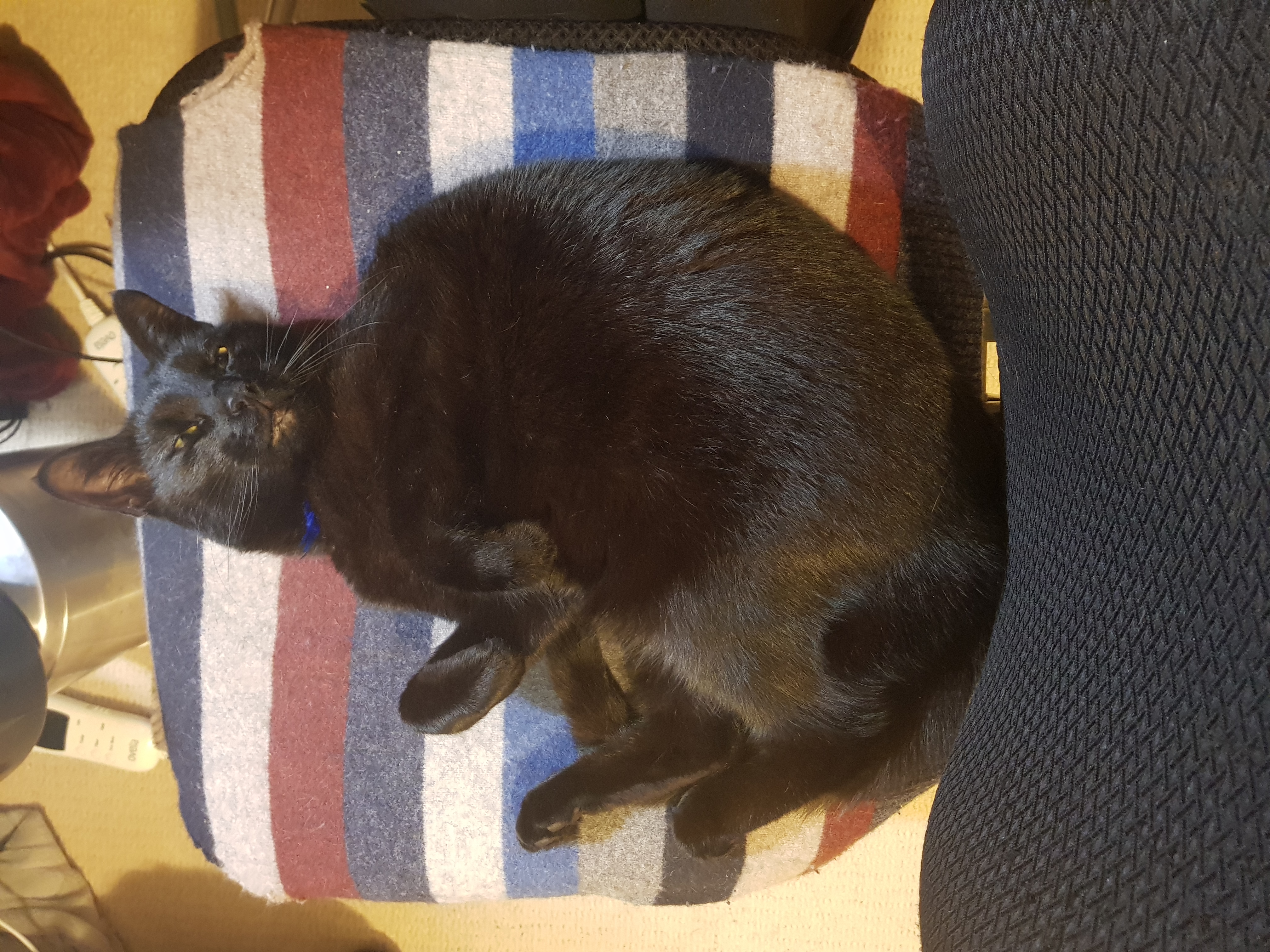}
    \hspace{10pt}
\includegraphics[width=0.4\textwidth, angle=270]{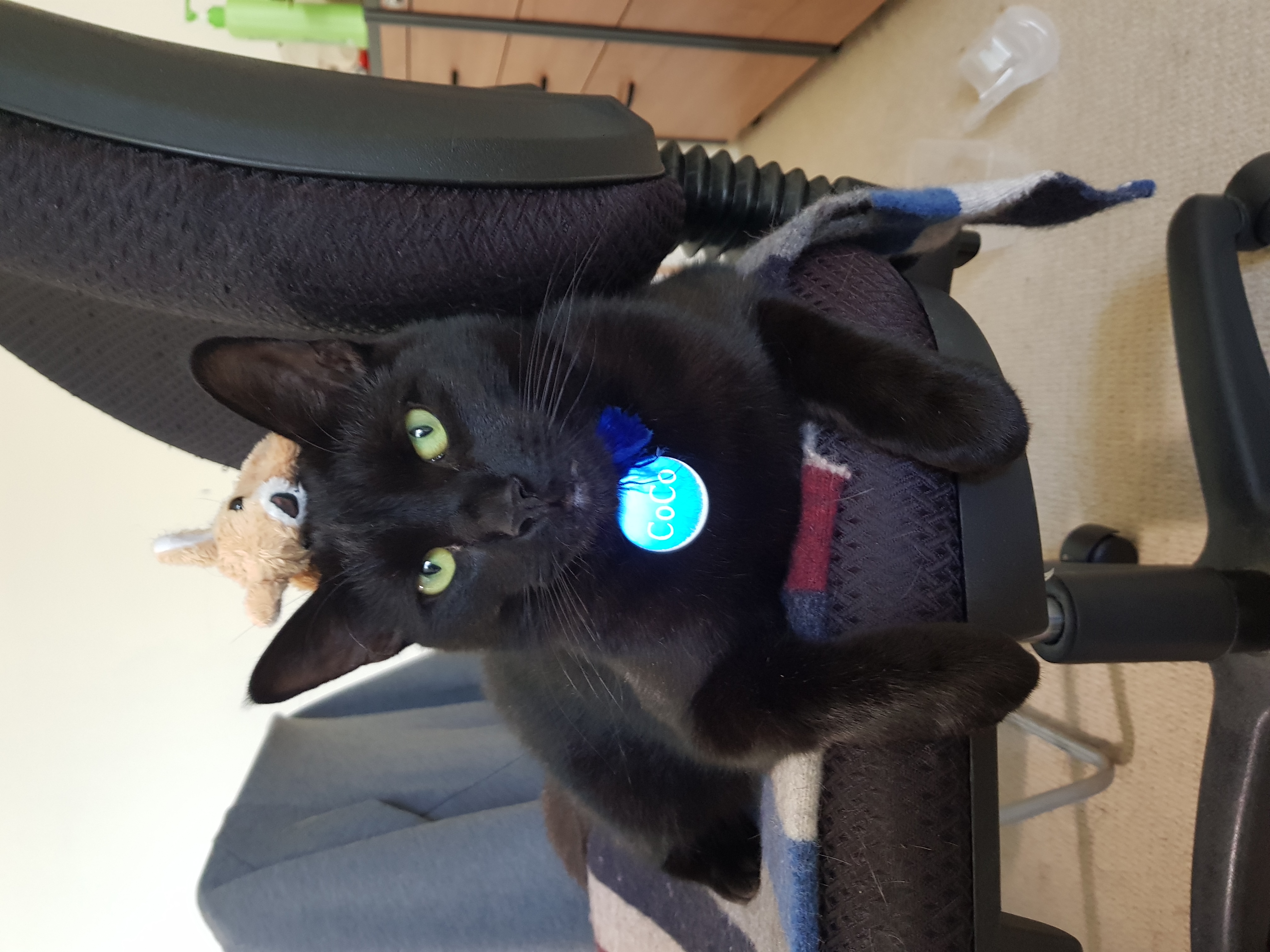}
    \caption{Source images being converted into histograms of pixel intensities.}\label{fig:coco}
\end{figure}

\begin{figure}
    \centering
    \includegraphics[width=1.0\textwidth]{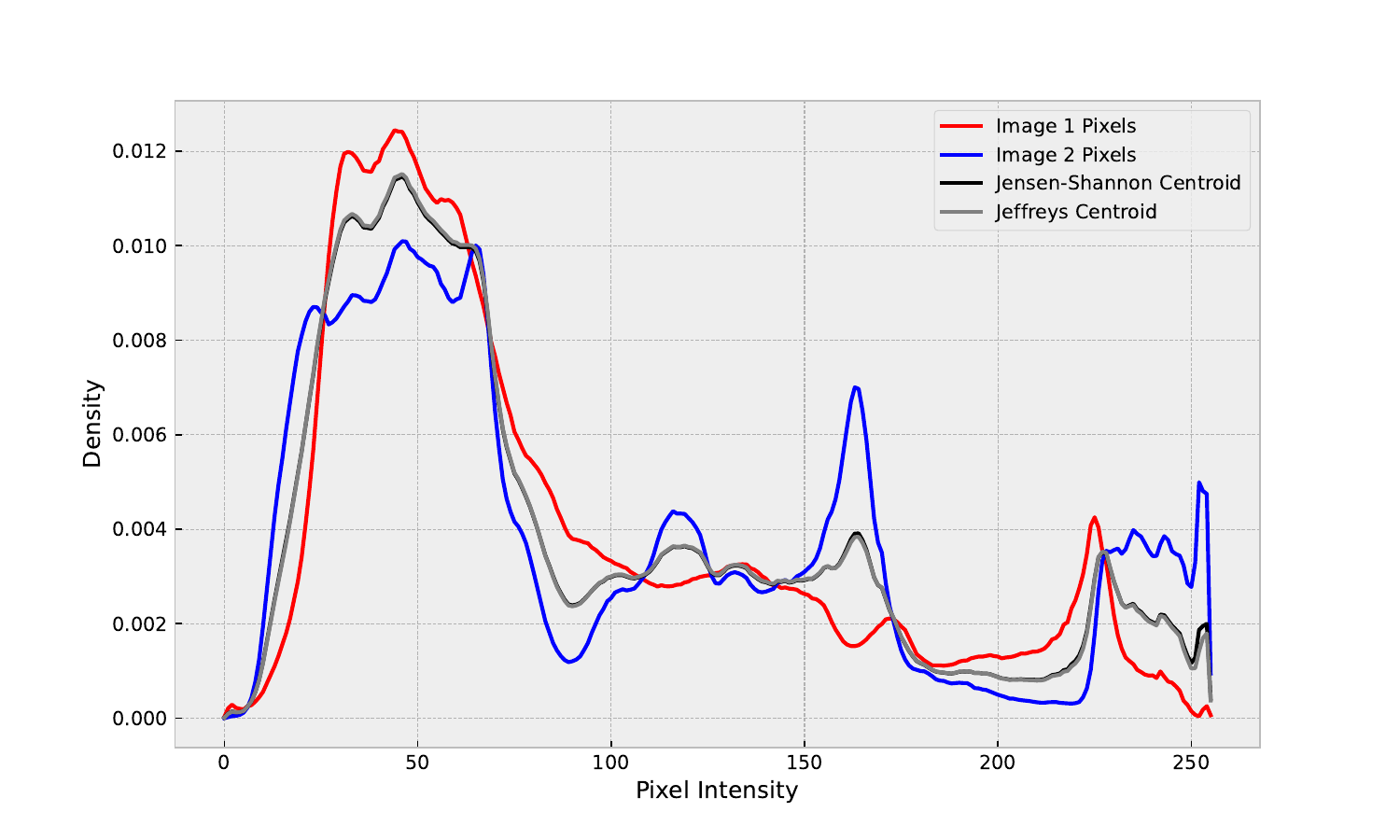}
    \caption{Jensen-Shannon and Jeffreys centroids of pixel intensity histograms of obtained from images displayed in Figure~\ref{fig:coco}.}\label{fig:hist}
\end{figure}

To demonstrate this, we consider the task of computing centroid on histograms
(categorical distributions) of images' pixel intensity distributions. We
specifically discretize the pixel intensity into 256 values corresponding to
the average value over the 3 RGB color channels.
The images that we consider in this example are those presented in Figure~\ref{fig:coco}.
The corresponding Jensen-Shannon and Jeffreys centroids are presented in Figure~\ref{fig:hist}.

\begin{lstlisting}
import requests
import matplotlib.pyplot as plt
import numpy as np
from PIL import Image


def image_gray_hist(image_url: str, ratio: float = 1.0) -> np.ndarray:
    # Get image from url
    image_path = requests.get(image_url, stream=True).raw
    image = Image.open(image_path)
    size = image.size
    image = image.resize((int(size[0] * ratio), int(size[1] * ratio)))
    im_array = np.array(image)

    # Discritize pixel intensity
    pixel_array = np.mean(im_array.reshape(-1, 3), axis=1)
    pixel_array = np.rint(pixel_array)

    # Make intensity array
    hist = np.zeros(256)
    for p in pixel_array:
        hist[int(p)] += 1

    # Ensures we are in the interiors of the simplex
    hist += 1e-8
    hist = hist / np.sum(hist)

    return hist


image_url_1 = "https://raw.githubusercontent.com/alexandersoen/pyBregMan/main/img/coco_1.jpg"
image_url_2 = "https://raw.githubusercontent.com/alexandersoen/pyBregMan/main/img/coco_2.jpg"

hist_1 = image_gray_hist(image_url_1)
hist_2 = image_gray_hist(image_url_2)

from bregman.application.distribution.exponential_family.categorical import \
    CategoricalManifold
from bregman.base import LAMBDA_COORDS, DualCoords, Point

cat_manifold = CategoricalManifold(k=256)

values = np.arange(cat_manifold.k)

cute_cat_1 = Point(LAMBDA_COORDS, hist_1)
cute_cat_2 = Point(LAMBDA_COORDS, hist_2)

# Let us work in the discrete mixture manifold space now.
mix_manifold = cat_manifold.to_discrete_mixture_manifold()
cute_mix_1 = cat_manifold.point_to_mixture_point(cute_cat_1)
cute_mix_2 = cat_manifold.point_to_mixture_point(cute_cat_2)


from bregman.barycenter.bregman import SkewBurbeaRaoBarycenter

# Define barycenter objects
js_centroid_obj = SkewBurbeaRaoBarycenter(
    mix_manifold, dcoords=DualCoords.THETA
)
jef_centroid_obj = SkewBurbeaRaoBarycenter(
    mix_manifold, dcoords=DualCoords.ETA
)

# Centroids can be calculated
js_centroid = js_centroid_obj([cute_mix_1, cute_mix_2])
jef_centroid = jef_centroid_obj([cute_mix_1, cute_mix_2])

# Convert from theta-/eta-parameterization back to histograms (lambda)
js_cat = mix_manifold.point_to_categorical_point(js_centroid)
jef_cat = mix_manifold.point_to_categorical_point(jef_centroid)

js_hist = cat_manifold.convert_coord(LAMBDA_COORDS, js_cat).data
jef_hist = cat_manifold.convert_coord(LAMBDA_COORDS, jef_cat).data


# Plot intensity histograms and centroids
with plt.style.context("bmh"):
    plt.plot(values, hist_1, c="red", label="Image 1 Pixels")
    plt.plot(values, hist_2, c="blue", label="Image 2 Pixels")
    plt.plot(values, js_hist, c="black", label="Jensen-Shannon Centroid")
    plt.plot(values, jef_hist, c="grey", label="Jeffreys Centroid")

    plt.xlabel("Pixel Intensity")
    plt.ylabel("Density")

    plt.legend()
    plt.show()
\end{lstlisting}

\subsection{Gaussian Distributions: Chernoff Point / Information}
\label{subsec:chernoff-example}

\begin{figure}
    \centering
    \includegraphics[width=1.0\textwidth]{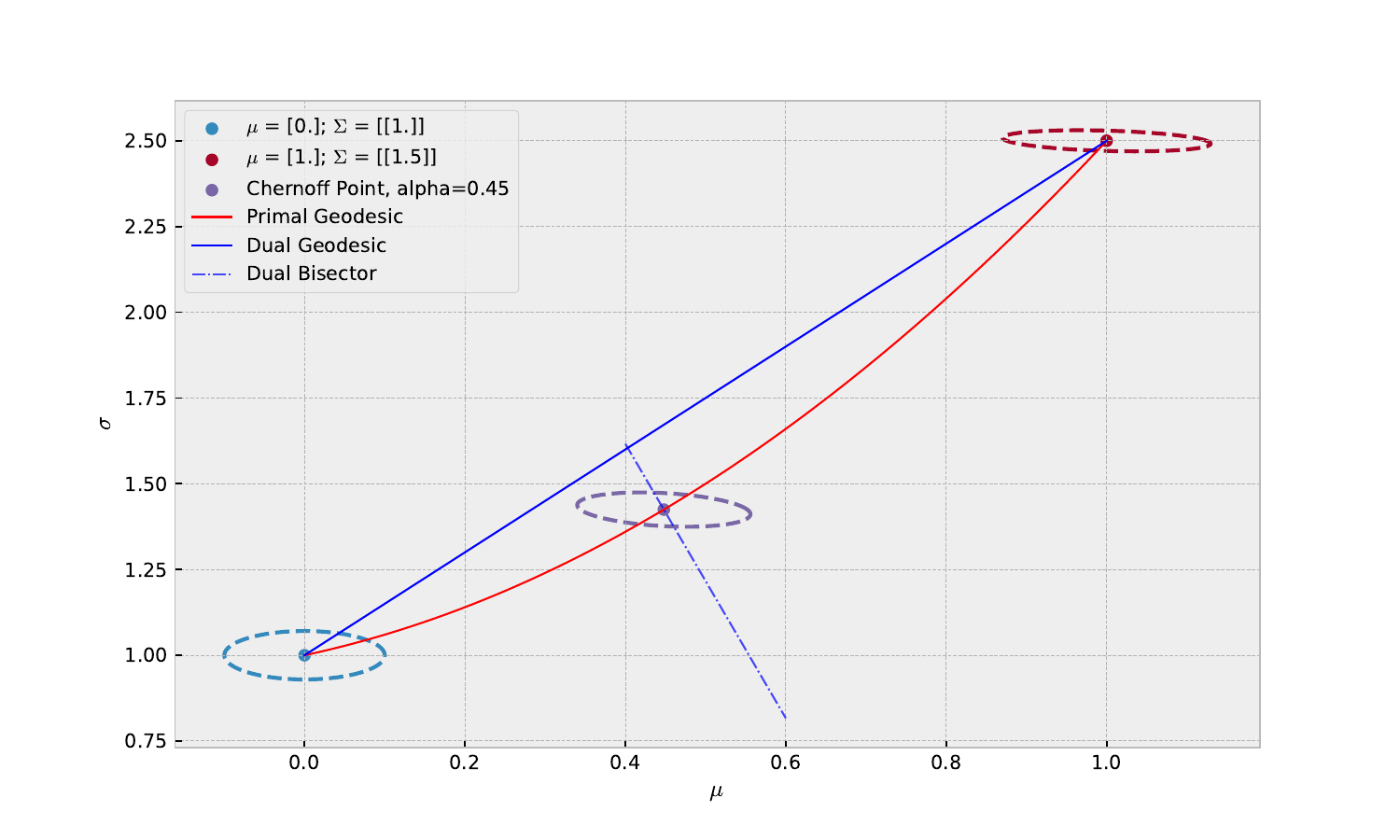}
    \caption{Calculation of the Chernoff point in the manifold of normal distributions (1D Gaussian manifold).}
    \label{fig:chernoff}
\end{figure}

The Chernoff distance~\cite{Chernoff-2022} or Chernoff information between two statistical distributions with densities $p(x)$ and $q(x)$ with respect to a base measure $\mu$ is defined by
$$
D_C(p:q)=\max_{\alpha\in (0,1)} -\log \int p^\alpha q^{1-\alpha} \mathrm{d}\mu.
$$
It can be shown that the optimal value $\alpha^*\in(0,1)$ can be obtained by intersecting a primal Bregman geodesic with a dual Bregman bisector~\cite{Chernoff-2022}.
Figure~\ref{fig:chernoff} displays the so called Chernoff point~\cite{Chernoff-2013} which can be calculated exactly on the univariate Gaussian manifold, and more generally approximated arbitrarily finely between any two densities of an exponential family. 

\begin{lstlisting}
import numpy as np

from bregman.application.distribution.exponential_family.gaussian import GaussianManifold
from bregman.base import ETA_COORDS, LAMBDA_COORDS, THETA_COORDS, DualCoords, Point
from bregman.dissimilarity.bregman import ChernoffInformation
from bregman.manifold.bisector import BregmanBisector
from bregman.manifold.geodesic import BregmanGeodesic
from bregman.visualizer.matplotlib import MatplotlibVisualizer, Visualize2DTissotIndicatrix

DISPLAY_TYPE = ETA_COORDS
VISUALIZE_INDEX = (0, 1)

# Define manifold + objects
manifold = GaussianManifold(1)

coord1 = Point(LAMBDA_COORDS, np.array([0.0, 1.0]))
coord2 = Point(LAMBDA_COORDS, np.array([1.0, 1.5]))

chernoff_point_alpha = ChernoffInformation(manifold).chernoff_point(coord1, coord2)
primal_geo = BregmanGeodesic(manifold, coord1, coord2, DualCoords.THETA)
dual_geo = BregmanGeodesic(manifold, coord1, coord2, DualCoords.ETA)

chernoff_point = primal_geo(1 - chernoff_point_alpha)

eta_bisector = BregmanBisector(
    manifold,
    coord1,
    coord2,
    dcoords=DualCoords.ETA,
)

theta_bisector = BregmanBisector(
    manifold,
    coord1,
    coord2,
    dcoords=DualCoords.THETA,
)

# Define visualizer
visualizer = MatplotlibVisualizer(manifold, VISUALIZE_INDEX, dim_names=(r"$\mu$", r"$\sigma$"))
metric_cb = Visualize2DTissotIndicatrix()

# Add objects to visualize
visualizer.plot_object(coord1, label=manifold.convert_to_display(coord1))
visualizer.plot_object(coord2, label=manifold.convert_to_display(coord2))
visualizer.plot_object(chernoff_point, label=f"Chernoff Point, alpha={chernoff_point_alpha:.2f}")
visualizer.plot_object(primal_geo, c="red", label="Primal Geodesic")
visualizer.plot_object(dual_geo, c="blue", label="Dual Geodesic")
visualizer.plot_object(eta_bisector, alpha=0.7, c="blue", label="Dual Bisector")

visualizer.add_callback(metric_cb)

visualizer.visualize(DISPLAY_TYPE)
\end{lstlisting}

\section{Extending \pyBregMan}

In the following, we go through an example of extending \pyBregMan by defining
a  new application manifold.
In particular, we will go through the process of defining a minimal 2D Bregman
manifold with primal generator $F$ specified as the extended negative Shannon
entropy function. Overall, the process can be broken down into two steps:

\begin{itemize}
\item The
first is to define the pair of Bregman generators used to define the Bregman
manifold. 
\item The second is to define quantities and methods associated with the
$\lambda$-coordinates, which is primarily used for displaying points on the
manifold.
\end{itemize}

Finally, we will present an example of implementing new manifold
specific geometric objects by defining a new Bregman ball object in this 2D
manifold with parameterized equation.

\subsection{Defining Generators using auto-differentiation}

The first and primary task we will consider is defining the generators for the
Bregman manifold.
We will utilize \VerbLB+AutoDiffGenerator+ from the
\VerbLB+bregman.manifold.generator+ submodule to define these Bregman generators.
By utilizing auto-differentiation, we will only be required to define the
generator functions itself and its derivatives will be automatically generated.
Although these derivatives can be automatically calculated, we will still be
required to implement the dual generator as there is no exact method for
automatically generating it.
As our primary generator, we utilize the extended negative Shannon entropy
function~\cite{MCIG-2019}:
\begin{equation}
    F(x)=F(x_1,x_2) = x_1 \log x_1 + x_2 \log x_2 - (x_1 + x_2).
\end{equation}

Through simple calculation, one can find that the dual generator:
\begin{equation}
    F^*(x) = \exp(x_1) + \exp(x_2).
\end{equation}

Defining these generators in \pyBregMan is straightforward:

\begin{lstlisting}
import autograd.numpy as anp
import numpy as np

from bregman.manifold.generator import AutoDiffGenerator

class EKL2DPrimalGenerator(AutoDiffGenerator):

    def _F(self, x: np.ndarray) -> np.ndarray:
        return anp.dot(x, anp.log(x)) - anp.sum(x)


class EKL2DDualGenerator(AutoDiffGenerator):

    def _F(self, x: np.ndarray) -> np.ndarray:
        return anp.sum(anp.exp(x))
\end{lstlisting}

One notable point in defining these generators in \pyBregMan is that we must
utilize `numpy` functions wrapped by the `autograd` library. Otherwise, the
auto-differentiation procedure will not work.

\subsection{Defining the Manifold}

With these two generators defined for the Bregman manifold, all we need to do
is define some quantities relating to a standard $\lambda$-parameterization.
This acts as the ``usual'' parameterization that is used to define objects on
the manifold. This can deviate from the canonical $\theta$-or
$\eta$-parameterizations. In our case, we will simply specify the
$\lambda$-parameterization to be equivalent to the $\theta$-parameterization.

In terms of \pyBregMan classes, we need to additionally define a corresponding
\VerbLB+DisplayPoint+ class. This is utilized for pretty printing the object and
will correspond to the $\lambda$-parameterization of points. To change the
default behaviour, one can override the \VerbLB+convert_to_display+ method in
\VerbLB+ApplicationManifold+.
Nevertheless, the corresponding application manifold can be defined as:

\begin{lstlisting}[firstnumber=16]
from bregman.application.application import ApplicationManifold
from bregman.base import DisplayPoint, DualCoords, Point
from bregman.dissimilarity.bregman import BregmanDivergence


class EKL2DPoint(DisplayPoint):

    def display(self) -> str:
        d_str = r"$\mathrm{EKL}({}, {})$"
        return d_str.format(self.data[0], self.data[1])


class EKL2DManifold(ApplicationManifold[EKL2DPoint]):

    def __init__(self) -> None:
        F_gen = EKL2DPrimalGenerator()
        G_gen = EKL2DDualGenerator()

        super().__init__(
            theta_generator=F_gen,
            eta_generator=G_gen,
            display_factory_class=EKL2DPoint,
            dimension=2,
        )

        self.eta_generator = G_gen  # Fix typing

    def _lambda_to_theta(self, lamb: np.ndarray) -> np.ndarray:
        return lamb

    def _lambda_to_eta(self, lamb: np.ndarray) -> np.ndarray:
        return self._theta_to_eta(lamb)

    def _theta_to_lambda(self, theta: np.ndarray) -> np.ndarray:
        return theta

    def _eta_to_lambda(self, eta: np.ndarray) -> np.ndarray:
        return self._eta_to_theta(eta)
\end{lstlisting}

Notice that the conversion between $\theta$-and $\eta$-parameterization are
automatically defined through inheritance via the \VerbLB+_theta_to_eta+ and
\VerbLB+_eta_to_theta+ methods. As such, we only needed to define the conversions
\wrt the $\lambda$-coordinates.

The \VerbLB+ApplicationManifold+ abstract class is generic and accepts a
\VerbLB+DisplayPoint+ type. We recommend specifying this to make it specific to
the manifold's display point type.

\subsection{Defining Geometric Objects}

Now utilizing the \VerbLB+EKL2DManifold+ manifold we have just defined, we will
define a \VerbLB+Ball+ object specific to the manifold. We will utilize the
closed form parameterization that is available for this specific Bregman
manifold.

\begin{lstlisting}[firstnumber=54]
from bregman.ball.base import Ball
from bregman.ball.parameterized import KL2DBregmanBallCurve
from bregman.base import Curve


class EKL2DBregmanBall(Ball[EKL2DManifold]):

    def __init__(
        self,
        manifold: EKL2DManifold,
        center: Point,
        radius: float,
    ) -> None:
        dcoords = DualCoords.THETA
        super().__init__(manifold, center, radius, dcoords.value)

        self.bregman_divergence = BregmanDivergence(manifold, dcoords=dcoords)

    def is_in(self, other: Point) -> bool:
        return self.bregman_divergence(self.center, other).item() < self.radius

    def parametrized_curve(self) -> Curve:
        return KL2DBregmanBallCurve(self.center, self.radius)
\end{lstlisting}

Here we are utilizing \VerbLB+KL2DBregmanBallCurve+ which was not linked to a
specific \VerbLB+Ball+ implementation.

The \VerbLB+Ball+ abstract class is generic and to make an subclass specific to a
manifold type one can make the typing concrete. All geometric style objects
follow pattern, \eg, \VerbLB+Geodesic+, \VerbLB+Dissimilarity+, \etc.

Notice that extended KL sphere can be exactly rendered using the Lambert W function~\cite{nielsen2021geodesic}.

\section*{Notebook}

A tutorial notebook entitled ``Data Representations on the Bregman Manifold'' demonstrating some aspects of the library is available at \url{https://gram-blogposts.github.io/}.

\section*{Acknowledgments}
FN warmly thanks Richard Nock for many fruitful discussions on Bregman divergences and its use in machine learning.
AS graciously thanks Ella Xi Wang for advice and assistance for module packaging and document generation in Python. 
AS would also like to thank Coco for providing excellent images for our code examples.


\begin{thebibliography}{10}

\bibitem{IG-2016}
Shun-ichi Amari.
\newblock {\em Information Geometry and Its Applications}.
\newblock Applied Mathematical Sciences. Springer Japan, 2016.

\bibitem{axen2023manifolds}
Seth~D Axen, Mateusz Baran, Ronny Bergmann, and Krzysztof Rzecki.
\newblock Manifolds. jl: an extensible julia framework for data analysis on
  manifolds.
\newblock {\em ACM Transactions on Mathematical Software}, 49(4):1--23, 2023.

\bibitem{Yoke-1997}
Ole~E Barndorff-Nielsen and Peter~Edmund Jupp.
\newblock Yokes and symplectic structures.
\newblock {\em Journal of statistical planning and inference}, 63(2):133--146,
  1997.

\bibitem{FY-2020}
Mathieu Blondel, Andr{\'e}~FT Martins, and Vlad Niculae.
\newblock {Learning with Fenchel-Young losses}.
\newblock {\em Journal of Machine Learning Research}, 21(35):1--69, 2020.

\bibitem{BVD-2010}
Jean-Daniel Boissonnat, Frank Nielsen, and Richard Nock.
\newblock {Bregman Voronoi diagrams}.
\newblock {\em Discrete \& Computational Geometry}, 44:281--307, 2010.

\bibitem{Bregman-1967}
Lev~M. Bregman.
\newblock The relaxation method of finding the common point of convex sets and
  its application to the solution of problems in convex programming.
\newblock {\em USSR computational mathematics and mathematical physics},
  7(3):200--217, 1967.

\bibitem{BurbeaRao-1982}
Jacob Burbea and C~Radhakrishna Rao.
\newblock Entropy differential metric, distance and divergence measures in
  probability spaces: A unified approach.
\newblock {\em Journal of Multivariate Analysis}, 12(4):575--596, 1982.

\bibitem{Eguchi-1992}
Shinto Eguchi.
\newblock Geometry of minimum contrast.
\newblock {\em Hiroshima Mathematical Journal}, 22(3):631--647, 1992.

\bibitem{matplotlib}
J.~D. Hunter.
\newblock Matplotlib: A 2d graphics environment.
\newblock {\em Computing in Science \& Engineering}, 9(3):90--95, 2007.

\bibitem{Kobayashi-2023}
Shimpei Kobayashi.
\newblock {Geodesics of multivariate normal distributions and a Toda lattice
  type Lax pair}.
\newblock {\em Physica Scripta}, 98(11):115241, 2023.

\bibitem{IGGeomstats-2023}
Alice Le~Brigant, Jules Deschamps, Antoine Collas, and Nina Miolane.
\newblock {Parametric information geometry with the package Geomstats}.
\newblock {\em ACM Transactions on Mathematical Software}, 49(4):1--26, 2023.

\bibitem{Matumoto-1993}
Takao Matumoto et~al.
\newblock {Any statistical manifold has a contrast function—On the
  $C^3$-functions taking the minimum at the diagonal of the product manifold}.
\newblock {\em Hiroshima Math. J}, 23(2):327--332, 1993.

\bibitem{Mitchell-1988}
Ann~FS Mitchell.
\newblock Statistical manifolds of univariate elliptic distributions.
\newblock {\em International Statistical Review/Revue Internationale de
  Statistique}, pages 1--16, 1988.

\bibitem{singularDFS-2021}
Naomichi Nakajima and Toru Ohmoto.
\newblock The dually flat structure for singular models.
\newblock {\em Information Geometry}, 4(1):31--64, 2021.

\bibitem{nakamura2001algorithms}
Yoshimasa Nakamura.
\newblock Algorithms associated with arithmetic, geometric and harmonic means
  and integrable systems.
\newblock {\em Journal of computational and applied mathematics},
  131(1-2):161--174, 2001.

\bibitem{Nakamura-2001}
Yoshimasa Nakamura.
\newblock Algorithms associated with arithmetic, geometric and harmonic means
  and integrable systems.
\newblock {\em Journal of computational and applied mathematics},
  131(1-2):161--174, 2001.

\bibitem{Chernoff-2013}
Frank Nielsen.
\newblock {An information-geometric characterization of Chernoff information}.
\newblock {\em IEEE Signal Processing Letters}, 20(3):269--272, 2013.

\bibitem{JSCentroid-2020}
Frank Nielsen.
\newblock {On a generalization of the Jensen--Shannon divergence and the
  Jensen--Shannon centroid}.
\newblock {\em Entropy}, 22(2):221, 2020.

\bibitem{nielsen2021geodesic}
Frank Nielsen.
\newblock On geodesic triangles with right angles in a dually flat space.
\newblock In {\em Progress in Information Geometry: Theory and Applications},
  pages 153--190. Springer, 2021.

\bibitem{ManyIG-2022}
Frank Nielsen.
\newblock The many faces of information geometry.
\newblock {\em Not. Am. Math. Soc}, 69(1):36--45, 2022.

\bibitem{Chernoff-2022}
Frank Nielsen.
\newblock Revisiting chernoff information with likelihood ratio exponential
  families.
\newblock {\em Entropy}, 24(10):1400, 2022.

\bibitem{FY-2022}
Frank Nielsen.
\newblock {Statistical divergences between densities of truncated exponential
  families with nested supports: Duo Bregman and duo Jensen divergences}.
\newblock {\em Entropy}, 24(3):421, 2022.

\bibitem{QAC-2023}
Frank Nielsen.
\newblock {Beyond scalar quasi-arithmetic means: Quasi-arithmetic averages and
  quasi-arithmetic mixtures in information geometry}.
\newblock {\em arXiv preprint arXiv:2301.10980}, 2023.

\bibitem{FR-2023}
Frank Nielsen.
\newblock Fisher-rao and pullback hilbert cone distances on the multivariate
  gaussian manifold with applications to simplification and quantization of
  mixtures.
\newblock In {\em Topological, Algebraic and Geometric Learning Workshops
  2023}, pages 488--504. PMLR, 2023.

\bibitem{inductivemean-2023}
Frank Nielsen.
\newblock What is... an inductive mean?
\newblock {\em Notices of the American Mathematical Society}, 70(11), 2023.

\bibitem{FisherRao-Nielsen-2024}
Frank Nielsen.
\newblock {Approximation and bounding techniques for the Fisher-Rao distances}.
\newblock {\em arXiv preprint arXiv:2403.10089}, 2024.

\bibitem{Nielsen-2024}
Frank Nielsen.
\newblock Divergences induced by the cumulant and partition functions of
  exponential families and their deformations induced by comparative convexity.
\newblock {\em Entropy}, 26(3):193, 2024.

\bibitem{ZF-2024}
Frank Nielsen.
\newblock Divergences induced by the cumulant and partition functions of
  exponential families and their deformations induced by comparative convexity.
\newblock {\em Entropy}, 26(3):193, 2024.

\bibitem{BR-2011}
Frank Nielsen and Sylvain Boltz.
\newblock {The Burbea-Rao and Bhattacharyya centroids}.
\newblock {\em IEEE Transactions on Information Theory}, 57(8):5455--5466,
  2011.

\bibitem{EF-2009}
Frank Nielsen and Vincent Garcia.
\newblock {Statistical exponential families: A digest with flash cards}.
\newblock {\em arXiv preprint arXiv:0911.4863}, 2009.

\bibitem{MCIG-2019}
Frank Nielsen and Ga{\"e}tan Hadjeres.
\newblock {Monte Carlo information-geometric structures}.
\newblock {\em Geometric Structures of Information}, pages 69--103, 2019.

\bibitem{nielsen2008smallest}
Frank Nielsen and Richard Nock.
\newblock On the smallest enclosing information disk.
\newblock {\em Information Processing Letters}, 105(3):93--97, 2008.

\bibitem{nielsen2009sided}
Frank Nielsen and Richard Nock.
\newblock {Sided and symmetrized Bregman centroids}.
\newblock {\em IEEE transactions on Information Theory}, 55(6):2882--2904,
  2009.

\bibitem{BregmanVPT-2009}
Frank Nielsen, Paolo Piro, and Michel Barlaud.
\newblock Bregman vantage point trees for efficient nearest neighbor queries.
\newblock In {\em IEEE International Conference on Multimedia and Expo}, pages
  878--881. IEEE, 2009.

\bibitem{BBtree-2009}
Frank Nielsen, Paolo Piro, and Michel Barlaud.
\newblock {Tailored Bregman ball trees for effective nearest neighbors}.
\newblock In {\em Proceedings of the 25th European Workshop on Computational
  Geometry (EuroCG)}, pages 29--32, 2009.

\bibitem{SEBB-2005}
Richard Nock and Frank Nielsen.
\newblock {Fitting the smallest enclosing Bregman ball}.
\newblock In {\em European Conference on Machine Learning}, pages 649--656.
  Springer, 2005.

\bibitem{ohara2024doubly}
Atsumi Ohara, Hideyuki Ishi, and Takashi Tsuchiya.
\newblock Doubly autoparallel structure and curvature integrals: Applications
  to iteration complexity for solving convex programs.
\newblock {\em Information Geometry}, 7(Suppl 1):555--586, 2024.

\bibitem{ohara1996dualistic}
Atsumi Ohara, Nobuhide Suda, and Shun-ichi Amari.
\newblock Dualistic differential geometry of positive definite matrices and its
  applications to related problems.
\newblock {\em Linear Algebra and its Applications}, 247:31--53, 1996.

\bibitem{Rao-1945}
Calyampudi~Radhakrishna Rao.
\newblock Information and the accuracy attainable in the estimation of
  statistical parameters.
\newblock {\em Bulletin of the Calcutta Mathematical Society}, 37(3):81--91,
  1945.

\bibitem{Rockafellar-1967}
Ralph~Tyrrell Rockafellar.
\newblock {Conjugates and Legendre transforms of convex functions}.
\newblock {\em Canadian Journal of Mathematics}, 19:200--205, 1967.

\bibitem{Shima-2007}
Hirohiko Shima.
\newblock {\em {The geometry of Hessian structures}}.
\newblock World Scientific, 2007.

\bibitem{Zhang-2004}
Jun Zhang.
\newblock Divergence function, duality, and convex analysis.
\newblock {\em Neural computation}, 16(1):159--195, 2004.

\bibitem{Zhang-2014}
Jun Zhang.
\newblock Divergence functions and geometric structures they induce on a
  manifold.
\newblock In {\em Geometric Theory of Information}, pages 1--30. Springer,
  2014.

\end{thebibliography}

    \appendix

    \section{Full Class Diagram}

    \begin{sidewaysfigure}
    
        \centering
        \includegraphics[width=\textwidth]{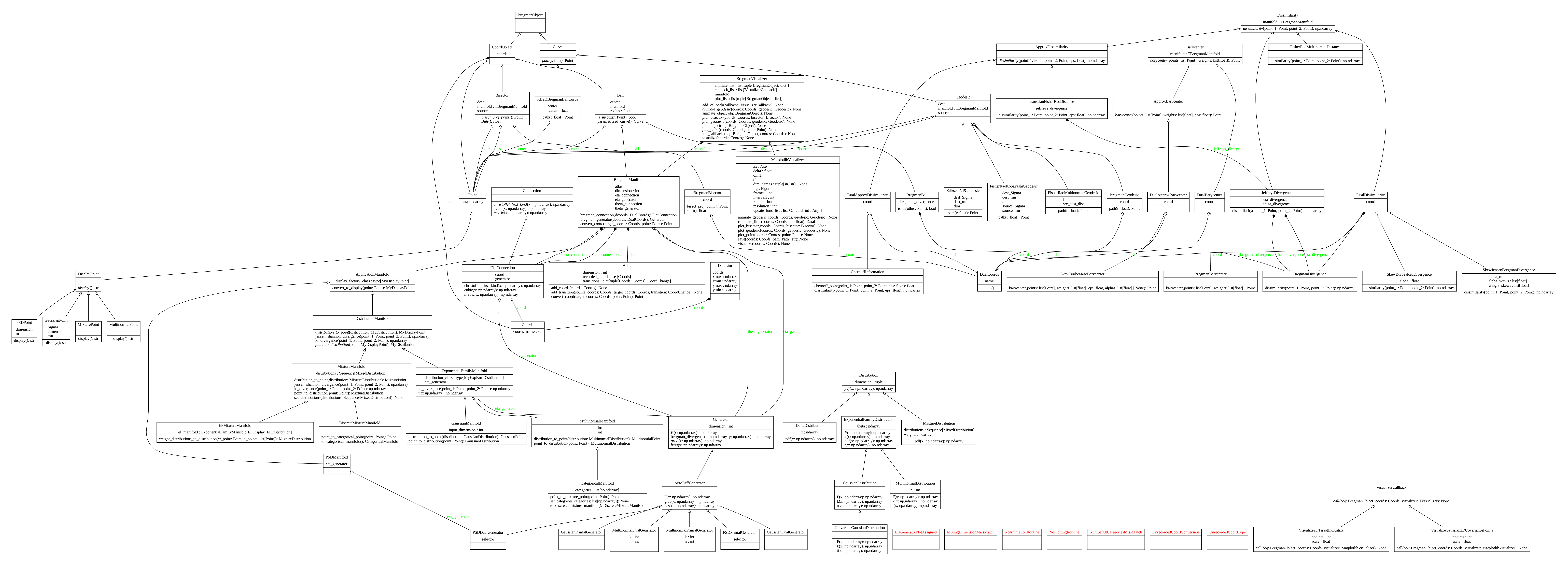}
        \caption{Full class diagram dependencies (zoom for details).}\label{fig:}
   
    \end{sidewaysfigure}

\end{document}